\documentclass[10pt,twocolumn,twoside]{IEEEtran}

\IEEEoverridecommandlockouts    
\usepackage{psfrag}
\usepackage{amsmath,amssymb,amsfonts,bbm,nicefrac,mathtools,lipsum}
\usepackage{latexsym}
\usepackage{graphicx}

\usepackage{caption}
\usepackage{subcaption}
\usepackage{balance}
\usepackage{colortbl}
\usepackage{fancyhdr}
\usepackage{epstopdf}
\usepackage{float}
\usepackage[hidelinks]{hyperref}
\usepackage{booktabs,tabularx}
\usepackage[noend]{algpseudocode}

\usepackage{soul}
\usepackage[ruled]{algorithm2e}

\begin{document}

\title{Safe Model Predictive Control Approach for Non-Holonomic Mobile Robots}

\author{Xinjie Liu, Vassil Atanassov
\thanks{Xinjie Liu and Vassil Atanassov are master students at Delft University of Technology, Netherlands. 
E-mail addresses: \{\texttt{X.LIU-47, v.v.atanassov}\}\texttt{@student.tudelft.nl}. \smallskip
}
}
\maketitle         

\begin{abstract}

We design an model predictive control (MPC) approach for planning and control of non-holonomic mobile robots. Linearizing the system dynamics around the reference trajectory gives a time-varying LQ MPC problem. We analytically show that by specially designing the MPC controller, the time-varying, linearized system can yield asymptotic stability around the origin in the tracking task. We further propose two obstacle avoidance methods. We show that by defining linearized constraint in \textit{velocity} space and explicitly \textit{coupling} the two control inputs based on current state, our second method directly accounts for the non-holonomic property of the system and therefore alleviates infeasibility of the optimization problems. Simulation results suggest that regarding both static and dynamic obstacle avoidance, the planned trajectories by our LQ MPC approach are comparably smooth and effective as solving non-linear programming (NLP) problems, but in a more efficient way. \footnote{The project video can be found on our \href{https://youtu.be/nYDxWkKvzZ8}{Youtube channel}}
\end{abstract}







\section{Introduction} \label{sec:intro}

\subsection{Introduction \& main contribution}

Non-holonomic mobile robots have a wide range of applications, such as differential drive cleaning and delivery robots, electric wheel chairs, and more. Model predictive control presents a promising way of achieving autonomous navigation for such robots in a dynamic environment. However, due to their non-linear dynamics and non-holonomic constraints, non-linear programming is typically employed, which is computationally expensive to solve and cannot have optimality guarantee. In this paper, we present a linear MPC solution to the tracking and obstacle avoidance problems for non-holonomic mobile robots.

Our main contributions are as follows:

\noindent (i) Introduced an MPC approach which works on linearised system dynamics and solves LQ MPC problems efficiently for robot trajectory tracking. 

\noindent (ii) Specially Designed terminal cost and terminal set, analytically showed local exponential stability for time-invariant sub-problem and recursive asymptotic stability for time-varying tracking problem. 

\noindent (iii) Proposed two methods of defining constraints in both \textit{position} and \textit{velocity} spaces for static and dynamic obstacle avoidance. We show that by explicitly coupling the two control inputs and considering the non-holonomic property of the system, our proposed constraint in velocity space can avoid infeasibility of the optimization problems and efficiently address obstacle avoidance problem. 

\subsection{Non-holonomic Mobile Robot}

One unicycle robot has nonlinear system dynamics with three states, where the kinematics can be described by:
\small
\begin{equation}
    \left(
    \begin{matrix}
    \dot x(t)\\
    \dot y(t)\\
    \dot \theta(t)\\
    \end{matrix}
    \right)
    = \dot z(t) = f(z(t), u(t)) = 
    \left(
    \begin{matrix}
    v(t)cos(\theta(t))\\
    v(t)sin(\theta(t))\\
    \omega(t) \\
    \end{matrix}
    \right)
\end{equation}
\normalsize
with an analytic vector field $f$: $\mathbb{R}^3 \times \mathbb{R}^2 \rightarrow \mathbb{R}^3$. The states $z = (x, y, \theta)^T$ $(m, m, rad)$ represent the x, y coordinates and the orientation of the robot. The control input is $u = (v, w)^T$ $(m/s, rad/s)$, where $v$ and $\omega$ are the linear and the angular speeds of the robot, respectively. The system is non-holonomic, meaning that it can not directly have lateral velocity, due to the non-slip constraints. 

By discretization (assuming piecewise constant control inputs on each sampling interval $[iT, (i+1)T], i \in \mathbb{N}_0$, with sampling time $T$ (seconds)), the discrete-time dynamics $f_{d, T}: \mathbb{R}^3 \times \mathbb{R}^2 \rightarrow \mathbb{R}^3$ are given by:
\small
\begin{equation}
    z(k+1) = f_{d, T}(z(k), u) =
    \left(
    \begin{matrix}
    x\\
    y\\
    \theta\\
    \end{matrix}
    \right)+
    \left(
    \begin{matrix}
    \frac{v}{\omega}(sin(\theta + T\omega) - sin(\theta))\\
    \frac{v}{\omega}(cos(\theta) - cos(\theta + T\omega))\\
    T\omega\\
    \end{matrix}
    \right)
\end{equation}
\normalsize


\section{Model predictive control design}

\subsection{Model Linearization}

We consider our tracking task as a setpoint stabilization problem, where the setpoint is time-varying and is moving along the reference trajectory $[x_{ref}(k), y_{ref}(k), \theta_{ref}(k)]^T$, where $k$ is a given time-step. Our MPC regulator steers the system to the generalized origin ($(\Tilde{x}, \Tilde{u})=(0, 0)$) and stabilizes the system. 

We linearise the system dynamics using first-order Taylor approximation along the reference trajectory and use Euler approximation for discretisation, which gives:

\small
\begin{equation*}
    e(k+1) = A(k)e(k) + Bu_b(k)
\end{equation*}
\normalsize
\footnotesize
\begin{equation}\label{model}
\begin{aligned}
    A(k) &=     
    \left[
    \begin{matrix}
    1 & \omega_r(k)T & 0\\
    -w_r(k)T & 1 & v_r(k)T\\
    0 & 0 & 1\\
    \end{matrix}
    \right],
    B &= 
    \left[
    \begin{matrix}
    -T & 0\\
    0 & 0\\
    0 & -T\\
    \end{matrix}
    \right]
\end{aligned}
\end{equation}
\normalsize

Note that at every timestep, $u_b$ is only feedback control term, we also use feed-forward term $u_f = [v_r, w_r]^T$ as a prior. For the details of the derivation, please refer to \cite{derivation}.

\subsection{Problem Formulation}
By linearizing the system dynamics around the reference trajectory $(x_{ref}, u_{ref})$, we have a \textit{Linear Time-Varying} (LTV) MPC problem. At each time step $i$:
\small
\begin{align*}
    \underset{u}{\text{min}} \hspace{0.2cm} V_N(x, u, i) &= \sum_{k=i}^{i+N-1} x(k)^TQx(k) + u(k)^TRu(k) \\
                       &+ V_f(x, i) \\
    \text{s.t.} \hspace{0.2cm} &x(k+1) = A(k)x(k) + Bu(k) + r_k\\
    & x(i+N) \in \mathbb{X}_{\textup{f}}(i)\\
    &|u(k)+u_{ref}(k)| \leq |u|_{max}\\
    & p(k) \in p_{free} \hspace{0.2cm} \forall k \in \{i,i+1,...,i+N\}
\end{align*}
\normalsize
where $V_f(x, i)$ is terminal cost, affine term $r_k$ is an offset, in case that the reference trajectory is not physically feasible for the system, otherwise this term is set to 0. The constraint $p(k) \in p_{free}$ is a collision avoidance constraint and will be discussed in the later sections. Note that in order to keep the notation consistent with literature, we write the error states $e(k)$ as generalized states $x(k)$.

We design the stage cost function as $\ell(x,u) = x(k)^TQx(k) + u(k)^TRu(k)$, the terminal cost function as $V_{\textup{f}}(x,i) = x(i+N)^TP(i)x(i+N)$, the terminal set as a sublevel set of the terminal cost $\mathbb{X}_{\textup{f}}(i) = \{x\in \mathbb{R}^3| x^TP(i)x \leq c\}$, where $c$ is a hyper-parameter, and $Q \succeq 0, R \succ 0$. The $P$ matrix is design in a way such that $P(i) = A_K(i)^T P(i+1) A_K(i) + Q_K(i)$, in order to ensure recursive stability, which will be discussed in the section \uppercase\expandafter{\romannumeral3}. Therefore, both the terminal cost function and the terminal set are time-varying.

\subsection{Obstacle Avoidance}

We propose two methods for obstacle avoidance: hyper-plane constraint in \textit{position space}  (only static obstacles) and \textit{velocity space} (both statics and dynamic obstacles).\\

\subsubsection{Motivation}

For avoiding obstacles using an MPC controller with hard constraints (as opposed to soft constraints added on top of the cost function, like repulsive fields\cite{potential_field}), one common practice is to limit the Euclidean distance from the robot to the obstacle, which requires quadratic constraints. However, because of the non-convex nature of the feasible area for the obstacle avoidance problem, non-linear programming (NLP) is normally employed, which is significantly more computationally expensive than solving quadratic programming (QP) problems, as in our case. Also, NLP cannot guarantee to find a global minimum, and therefore is sub-optimal. There are numerous ways of dealing with this, for example one approach is to use a convex approximation of the non-convex QCQP and solving it through Sequential Convex Programming \cite{SCP}. Another interesting method of overcoming this problem is by substituting the quadratic constraints with penalising quadratic costs - as seen in \cite{hermans2018penalty}.

Another kind of effective methods is to formalize the obstacle avoidance problem as a Mixed-Integer Linear Programming (MILP) problem. The Mixed-Integer approach uses a combination of linear constraints around the obstacle, and requires at least one of those constraints to be satisfied\cite{MILP}. But the MILP is also more expensive to be solved in real-time compared to QP. Given this, we propose two methods to define linear constraints for obstacle avoidance, which results in QPs and is therefore much more computationally cheaper.\\

\subsubsection{Method 1: hyper-plane in position space}
\begin{figure}
    \centering
    \includegraphics[scale=0.3]{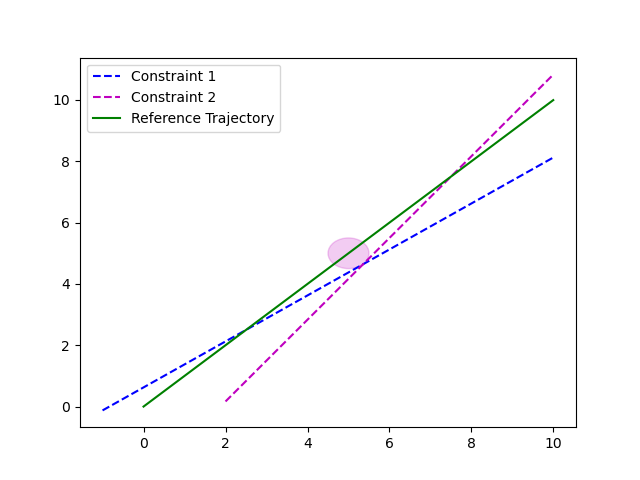}
    \caption{Position space constraint. One of the two half-planes is active, depending on the relative position of the robot.}
    \label{fig:obstacle}
\end{figure}

In our approach, only one constraint is active at a given time step, however, the constraint itself changes depending on the relative position of the robot w.r.t the obstacle. As Fig.\ref{fig:obstacle} suggests, two possible variations of the constraints are imposed - one for when the robot is in front of the obstacle, and one for when it is behind it. The two hyper-planes are defined by finding the unit vector between the robot and the obstacle and then rotating it by a safety angle $\theta$. This constraint divides the space in two half-planes and constrains the robot's planned position to the within the one without the obstacle.\\
Thus the constraints become of the form:
\begin{equation}
    \vec{n} \vec{x} \leq a
\end{equation}
where $\vec{n}$ is the vector orthogonal to the $\theta$-rotated hyper-plane, and $a$ is the offset of the constraint.
These constraints can then be formulated as polytopic constraints in the error space, namely by substituting $\vec{x} = \vec{e} + \vec{x}_{ref}(i)$ for each time step.\\

\subsubsection{Method 2: hyper-plane in velocity space}


Constraints in position space are most intuitive and direct. However, they do not account for the non-holonomic property of the system and therefore can lead to infeasible optimization problems. Adding slack variables to loose the constraints helps but can lose the collision avoidance guarantee.

To tackle this problem and generalize the collision avoidance, we propose to explicitly account for the non-holonomic constraint by coupling the two decision variables, which is done by defining feasible area in \textit{velocity space} instead of in \textit{position space}. 

Our method is inspired by \textit{Velocity Obstacles}\cite{ORCA} and \textit{Dynamic Window}\cite{DW} obstacle avoidance methods. Here we assume there are two identical robots. As suggested by Fiorini et al.(1998)\cite{VO}, a static obstacle in \textit{velocity space} can be described as an area consisting of infinite circles, with the center moving along the line between the robot and the obstacle and the radius varying with the movement:
\small
\begin{equation*}
    VO^{\tau} = \mathop{\cup} \limits_{0 \leq t \leq \tau} D(\frac{\vec{p}_j - \vec{p}_i}{t}, \frac{r_i+r_j}{t})
\end{equation*}
\normalsize
where $D(x, r)$ is the disc with center $x$ and radius $r$, $\tau$ is planned time horizon, $\vec{p_j}$ and $\vec{p_i}$ are the positions of objects j and i, respectively. A moving obstacle with velocity \textbf{$\vec{v_j}$} will translate the origin in velocity space:
\small
\begin{equation*}
    VO^{\tau} = \mathop{\cup} \limits_{0 \leq t \leq \tau} D(\frac{\vec{p_j} - \vec{p_i}}{t}+\vec{v_j}, \frac{r_i+r_j}{t})
\end{equation*}
\normalsize

\begin{figure}
    \centering
    \includegraphics[scale=0.27]{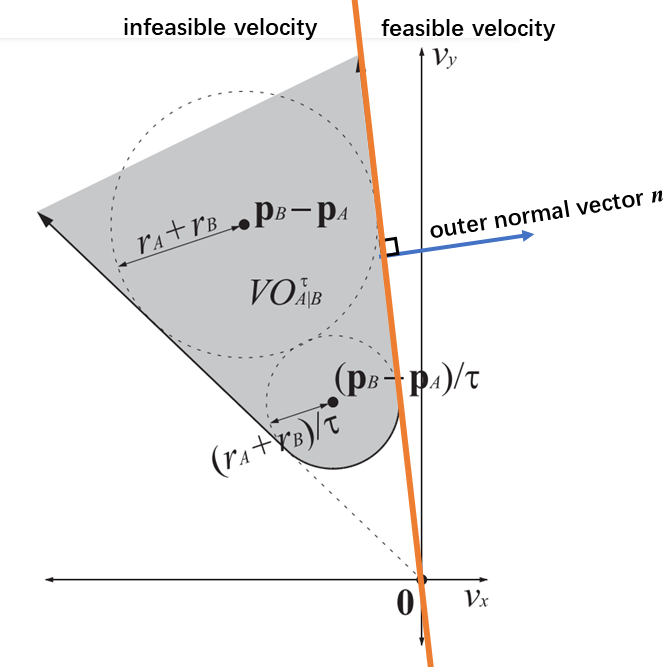}
    \caption{ Velocity space partition. \textbf{$\vec{p}_A$} is the position of the robot. Orange hyper-plane splits the velocity space. Dynamic obstacle will translate the origin. We generalize this idea to non-holonomic systems. Adapted from from \cite{VO}.}
    \label{fig:VO}
\end{figure}


Since the resulting feasible area is non-convex we impose a hyper-plane that is tangent to the velocity obstacle. As can be seen, this constraint leads to a conservative, but feasible area. When generalizing to cooperative collision avoidance of multiple-robots, they can each take solutions on the respective side of the constraint. 

Alonso-Mora et al. (2013) \cite{ORCA} assumed the robots can change their velocity instantaneously both in \textit{direction} and in \textit{magnitude}, and gave an error tolerance for non-holonomic robots, while we only assume the robots can change their velocity instantaneously in \textit{magnitude} and \textit{explicitly} account for the non-holonomic constraint in our formulation, which gives a more theoretically sound constraint. 

As Fig.\ref{fig:VO} shows, first we find an outer normal vector $\vec{n}$, which is perpendicular to the hyper-plane constraint in velocity space, pointing in the direction of the feasible velocity area. For a single robot obstacle avoidance, assume at any time step $i$, our robot has state $[x, y, \theta]^T$. By giving control inputs $[v,\omega ]^T$,for the next time-step $i+1$, we have its velocity vector \textbf{$\vec{u}$}:
\small
\begin{equation*}
    \vec{u} = 
    \left[
    \begin{matrix}
    vcos(\hat{\theta})\\
    vsin(\hat{\theta})\\
    \end{matrix}
    \right] =
    \left[
    \begin{matrix}
    (u_r+e_u)cos(\theta + \omega dt)\\
    (u_r+e_u)sin(\theta + \omega dt)\\
    \end{matrix}
    \right]
\end{equation*}
\normalsize
where $[u_r, \omega_r]^T$ are reference control inputs for tracking the reference trajectory, $e_u = \|u\|_2 - u_r$ is error between the planned velocity and the reference. The $\ell 2$ norm on the planned velocity means we can only control its magnitude directly. Note that we omit the transformation to the robot local frame here for simplifying the derivation, but the results are general.

The idea here is to restrict the velocity vector at the next time-step around the current one by using an Euler approximation to the heading angle, which takes into account the non-holonomic property of the robot. 

For a obstacle with velocity \textbf{$\vec{v}_j$}, we take the dot product between its velocity and the normal vector \textbf{$\vec{n}$}: $a = \vec{n}\vec{v}_j$. 

For the planned velocity vector \textbf{$u$} that is collision-free, its projection onto the normal vector $n$ should be greater than the obstacle's velocity projection $a$, since $v_j$ is exactly on the hyper-plane that we defined:
\small
\begin{equation*}
    \vec{n}^T
    \left[
    \begin{matrix}
    (u_r+e_u)cos(\theta + \omega dt)\\
    (u_r+e_u)sin(\theta + \omega dt)\\
    \end{matrix}
    \right] \geq a
\end{equation*}
\normalsize
which gives our ideal constraint on the control inputs $[\|u\|_2, \omega]^T$:
\small
\begin{equation}
    f(\|u\|_2, \omega) = n_1\|u\|_2cos(\theta + \omega dt) + n_2 \|u\|_2 sin(\theta + \omega dt) -a \geq 0
\end{equation}
\normalsize
Equation (5) is non-linear, which cannot be used by our QP problem. We linearize $f(\|u\|_2, \omega)$ around the reference control inputs $[u_r, \omega_r]$ using first-order Taylor approximation:
\small
\begin{equation}
    \begin{split}
        &f(\|u\|_2, \omega)|_{(u_r, \omega_r)} \approx f(u_r, \omega_r)\\ 
        &+ \frac{\partial f(\|u\|_2, \omega)}{\partial \|u\|_2}|_{(u_r, \omega_r)}(\|u\|_2 - u_r)\\ 
        &+ \frac{\partial f(\|u\|_2, \omega)}{\partial \omega}|_{(u_r, \omega_r)}(\omega - \omega_r)
    \end{split}
\end{equation}
\normalsize
which gives:
\small
\begin{equation}
    \begin{split}
        &-n_1u_rcos(\theta+\omega_r dt)-n_2u_rsin(\theta+\omega_r dt)\\
        &-n_1cos(\theta+\omega_r dt)e_u - n_2sin(\theta+\omega_r dt)e_u\\
        &+n_1u_rsin(\theta+\omega_r dt)dte_w - n_2u_rcos(\theta+\omega_r dt)dte_\omega+a\leq 0
    \end{split}
\end{equation}
\normalsize
This is the final constraint that we give to the solver, where $[e_u, e_{\omega}]^T$ are decision variables. As suggested above, our method can deal with both \textit{static} and \textit{dynamic} obstacles. Note that this method is general and can be used for multiple robots coordination.

\section{Asymptotic stability}




In this section, we first consider a simplified scenario where our MPC controller can be viewed as a linear time-invariant MPC regulator, and therefore local exponential stability can be proved. We further modify our concepts and assumptions for generalizing to the time-varying tracking case.

\subsection{Linear time-invariant MPC regulator}

For our linearized error dynamics Eq.\ref{model}, a key insight is that when the reference control inputs $\{v_r, w_r\}$ are constant, the model itself is therefore time-invariant. Since the reference trajectory is still time-varying, we further simplify the scenario to regulating the robot to a fixed, controllable point. Under the assumption of the robot is close enough to the reference trajectory so that our linearization still holds, we could instead analyse closed-loop LTI system stability.

For the linearized system dynamics Eq.\ref{model}, \normalsize the controllability matrix $[B, AB, A^2B, ..., A^NB]$ is full row rank, as long as the reference control inputs $\{v_r, w_r\}$ are not both zero. That is, the system is controllable, once it starts to move \cite{linearMPC}.

As discussed in Section 2.5 of \cite{textbook}, for some continuous linear system dynamics $f$, quadratic terminal cost $V_f$, stage cost $\ell(x,u)$ and terminal state constraint $\mathbb{X}_f$, if assumptions 2.2, 2.3, 2.14, 2.17 are satisfied, according to theorem 2.19 (2.21), the origin is asymptotically (exponentially) stable. Now we show that in our case, these assumptions are satisfied.\\

\noindent \textit{Assumption 2.2 (Continuity of system and cost)} is satisfied since the system dynamics $f (\mathbb{Z}\rightarrow \mathbb{X})$, stage cost $\ell(x, u) (\mathbb{Z}\rightarrow\mathbb{R}_{\geq 0})$, terminal cost $V_f(x_N) (\mathbb{X}_f \rightarrow \mathbb{R}_{\geq 0})$ are continuous, and $f(0,0)=0$, $\ell(0,0)=0$ and $V_f(0)=0$. \\

\noindent \textit{Assumption 2.3 (Properties of constraint sets)} is satisfied since the set $\mathbb{Z}$ is closed and the set $\mathbb{X}_f \in \mathbb{X}$ is compact. As we showed above, the origin $(\Tilde{x}, \Tilde{u})=(0, 0)$ is within the set $\mathbb{X}_f$ and is an equilibrium of the system.\\

\noindent \textit{Assumption 2.14 (Basic stability assumption)} is satisfied in this case. For a single point regulation problem, we use the solution of the Riccati equation as the terminal cost $V_f(x) = \frac{1}{2}x^TPx$, which means we approximate the optimal cost-to-go for a constrained system with the infinite-horizon optimal cost-to-go $V^{uc}_{\infty}(x)$ for the corresponding unconstrained system. By at least using an LQR control gain around the origin, we can have Lyapunov decrease of the terminal cost (note that we use notation $(x,u)$ here in order to keep consistency with the reference):
\small
\begin{equation*}
    V_f(A_kx) = \frac{1}{2}(A_kx)^TP(A_kx) = \frac{1}{2}x^TA_k^TPA_k x
\end{equation*}
\normalsize
\noindent where 
\small
\begin{equation*}
    P = A_k^TPA_k + Q_k, A_k = A + BK, Q_k = Q + K^TRK
\end{equation*}
\normalsize
\noindent therefore:
\small
\begin{equation*}
    V_f(A_kx) = \frac{1}{2}x^TPx - \frac{1}{2}x^TQ_kx = V_f(x) - \ell(x,u)
\end{equation*}
\normalsize
Note that the only assumption for using the LQR control gain to yield Lyapunov decrease is $u=Kx \in \mathbb{U}$. We need to choose the $\mathbb{X}_f$, so that inside the terminal set, the input constraints are inactive. But this is a design choice, we define the terminal set as a sublevel set of terminal cost $\mathbb{X}_f:= \{x\in \mathbb{R}^n | V_f(x) \leq c \}$, and choose suitable $c$ such that all points inside satisfy the constraints. By Lyapunov's Invariant Set Theorem, any sublevel set of a Lyapunov function is in itself invariant - which also implies that any sublevel set of a CLF is in itself control invariant, thus the condition for $\mathbb{X}_f$ is satisfied.

Furthermore, for assumption 2.14 (b), we have:
\small
\begin{equation}
\begin{aligned}
    \ell(x,u) &= \frac{1}{2} (x^TQx+u^TRu) \geq \frac{1}{2} x^TQx \\
    &\geq \lambda_{min}(Q)|x|^2 = \alpha_1(|x|)
\end{aligned}
\end{equation}
\begin{equation}
    V_f(x) = \frac{1}{2}x^TPx \leq \frac{1}{2}\lambda_{max}(P)|x|^2 = \alpha_2(|x|)
\end{equation}
\normalsize
where $\lambda(\cdot)$ is the eigenvalue of the respective matrix. Since both $\alpha_1$ and $\alpha_2$ are linear functions of the norm of x, then it can be seen that they are also $K_{\infty}$ functions.\\

\noindent \textit{Assumption 2.17 (Weak controllability)} is satisfied as it is a weaker assumption than the controllability assumption.\\

Therefore, for the LTI MPC regulator, equilibrium $(\Tilde{x}, \Tilde{u}) = (0, 0)$ is exponentially stable. This implies that under the assumption that our linearization still holds, set point regulation problem always yields a stable closed-loop system.


\subsection{Linear time-varying MPC tracker}

For generalizing the asymptotic stability to time-varying case, according to \textit{Theorem 2.39 (Asymptotic stability of the origin: time-varying MPC)}\cite{textbook}, Assumptions 2.25, 2.26, 2.33, and 2.37 need to be checked. We now show that by specially designing the terminal cost matrix $P$ and the terminal set $\mathbb{X}_f$, the satisfaction of these assumptions can be derived analytically.\\

\noindent \textit{Assumption 2.25 (Continuity of system and cost; time -varying case)} and \textit{Assumption 2.26 (Properties of constraint sets; time-varying case)} are satisfied. The assumptions for properties of sets $\mathbb{X}, \mathbb{X}_f, \mathbb{U}, \mathbb{Z}$ and functions $\ell(x, u), V_f(x), f(x, u)$ discussed in \textit{Section A} can be generalized to time-varying case without loss of generality. Furthermore, the sets $\mathbb{U}(i), i\in\mathbb{I}_{\geq 0}$ are uniformly bounded by the input constraint set $\Tilde{\mathbb{U}}$, which is compact and time-invariant.\\

\noindent \textit{Assumption 2.33 (Basic stability assumption; time-varying case)}.
By careful selection of the terminal cost matrix $P(i)$ for each step $i$, we can ensure the Lyapunov cost decrease in Assumption 2.33 (a). Our choice is the following:
\small
\begin{equation}
    P(i) = A_K(i)^T P(i+1) A_K(i) + Q_K(i)
    \label{eq:terminal_cost_LTV}
\end{equation}
\normalsize
where $A_K(i)$ is the closed loop system $A_K(i) = A(i) +  B(i)K(i)$, and $Q_K(i) = Q(i) + K(i)^T R(i) K(i)$; where $K(i)$ comes from the solution of the DARE for the linearised dynamics of timestep $i=\{0,1,2,...T\}$. In our case the penalising terms Q and R are constant, which further simplifies the solution without loss of generality. The terminal cost matrix $P(i)$ for each step is calculated recursively backwards, starting from $P(T)$, which is the solution to the DARE at the final time step $T$.
Following Eq. \ref{eq:terminal_cost_LTV}, we can define the terminal cost as 
\small
\begin{equation}
  V_f(f(x,u),i+1) = \frac{1}{2}x^T A_K(i)^T P(i+1) A_K(i) x 
  \label{eq:terminal_cost_2}
\end{equation}
\normalsize
Rearranging Eq. \ref{eq:terminal_cost_LTV} and substituting for $ A_K(i)^T P(i+1) A_K(i)$ in Eq. \ref{eq:terminal_cost_2}, we obtain the following formulation:
\small
\begin{equation}
\begin{aligned}
    V_f(f(x,u),i+1) & =  \frac{1}{2}x^T (P(i) - Q_K(i)) x \\
    & = \frac{1}{2}x^T P(i) x - \frac{1}{2}x^T Q_K(i) x \\
    & = V_f(x,i) - \frac{1}{2}x^T Q_K(i) x 
\end{aligned}
\end{equation}
\normalsize
Following our definition of $Q_K(i) = Q(i) + K(i)^T R(i) K(i)$ and since $u = K x$, then  $\frac{1}{2}x^T Q_K(i) x$ is exactly the stage-cost $\ell(x,u,i)$, and therefore we have proved Lyapunov decrease of Assumption 2.33 (a).
\small
\begin{equation}
     V_f(f(x,u),i+1) \leq  V_f(x,i) - \ell(x,u,i)
\end{equation}
\normalsize
Furthermore, for our design of terminal set $\mathbb{X}_f(i) = \{x\in \mathbb{R}^3|x^TP(i)x \leq c\}$, at any timestep i, if state $x(i) \in \mathbb{X}_f(i)$, then at the next timestep we have:
\small
\begin{equation}
    \begin{split}
        &x(i+1)^TP(i+1)x(i+1) = x(i)^T A_K(i)^T P(i+1) A_K(i) x(i)\\ 
        &= x(i)^TP(i)x(i) - x(i)^TQ_k(i)x(i) \leq c
    \end{split}
\end{equation}
\normalsize
That is, by decreasing of the terminal cost, our terminal set $\mathbb{X}_f$ is a \textit{sequentially control invariant} set.\\

For \textit{Assumption 2.33 (b)}, we have:
\small
\begin{equation}
    \begin{split}
        &\ell(x,u,i) = \frac{1}{2} (x^TQx+u^TRu) \geq \frac{1}{2} x^TQx\\ &\geq \lambda_{min}(Q)|x|^2 = \alpha_1(|x|)\\
        &V_f(x, i) = \frac{1}{2}x^TP(i)x \leq \frac{1}{2}\lambda_{max}(P(j))|x|^2 = \alpha_2(|x|)
    \end{split}
\end{equation}
\normalsize
where $\lambda_{max}(P(j))$ is the largest eigenvalue among all the $P(j)$ matrices. Therefore, we have showed the satisfaction of \textit{Assumption 2.33}.\\

\noindent \textit{Assumption 2.37 (Uniform weak controllability)} is satisfied. We can easily find a $\mathcal{K}_{\infty}$ upper bound for the optimal value function, such that:
\small
\begin{equation}
    \begin{split}
    V_N^o(x,i) &\leq \sum_{k=i}^{i+N-1}\frac{1}{2}x(k)^T\lambda_{max}(Q)x(k)\\
    &+\frac{1}{2}N|u|_{max}^T\lambda_{max}(R)|u|_{max}\\
    &+\frac{1}{2}x(i+N)^T\lambda_{max}(P(j))x(i+N)
    \end{split} 
\end{equation}
\normalsize
By substituting all the $x(k), k>i$, we yield the upper bound function $\alpha(|x|)$.To summarize, by specially designing the terminal cost matrices $P(i)$ and the terminal set $\mathbb{X}_f$, we analytically showed that the time-varying MPC tracker yields recursive asymptotically stability around the equilibria along the reference trajectory and control inputs $[x_{ref}, u_{ref}]^T$.

\subsection{Obstacle avoidance}

For the case of obstacle avoidance, since the robot is deviating from the reference and our assumptions are no longer guaranteed to hold, we numerically show that our algorithm can converge well back to the reference after avoiding the collision. An extension for re-gaining the assumptions can be iteratively re-plan the reference and re-linearize the system dynamics. This method is called iterative LQR (iLQR) \cite{iLQR}.

\subsection{Terminal Set}

As mentioned before, our design of terminal set is a sublevel set of the terminal cost $\mathbb{X}_f(i) = \{x | V_f(x,i) \leq c \}$ and it is a control invariant set. We now show how we choose the parameter $c$, so that all the points within the set also satisfy the states and inputs constraints.

Our terminal set is an 3D ellipsoid, whose semi-axes are defined by the three eigenvectors of $P(i)$. We (outer) approximate $\mathbb{X}_f(i)$ by a convex polyhedron (in this case a hexahedron) around the ellipse. Due to the convexity of the polyhedral set, if we show that each of the 8 vertices of the polyhedron lies within the feasible set, then the ellipsoid constrained by the polyhedron is also guaranteed to be in the feasible set. Numerically, our implementation can be described by the pseudo-code in Algorithm 1.

\begin{algorithm}
\caption{Terminal set calculation}
 Initialisation\\
 polyhedrons = [] \\
 c\_list = [] \\
 \For{every $P(i)$ in $P$}{
  $c$ = 10   \textit{// Start with an arbitrarily high $c$}\\
  feasible = False\\
  \small
  \While{not feasible}{
    $c$ = $c$ / 1.01  \textit{// Decrease $c$ by a small amount} \\
    vertices, feasible = calculate\_Terminal\_set($c, P(i)$)}
    polyhedrons[i] = vertices\\
    c\_list[i] = c}
return polyhedrons, c\_list
\end{algorithm}
\normalsize

\begin{figure}
    \centering
    \includegraphics[scale=0.4]{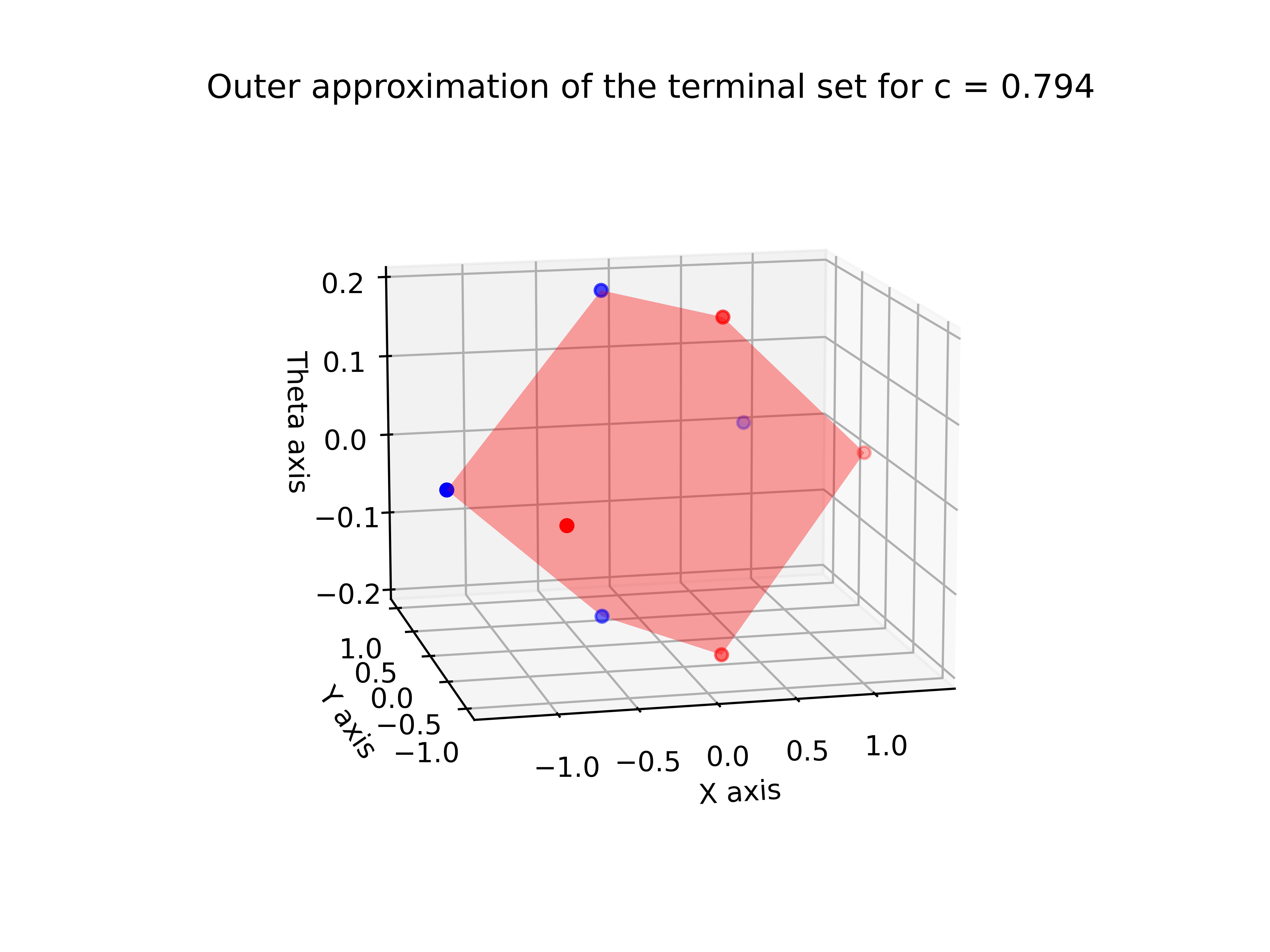}
    \caption{Outer approximation of the terminal set (at i=10) for the periodic sinusoidal trajectory, c = 0.794}
    \label{fig:terminal_set}
\end{figure}

Where the \textit{calculate\_Terminal\_set(c, P(i))} function fits a polyhedron based on $\lambda*c$, where $\lambda$ are the eigenvalues of $P(i)$; and then checks if each of the 8 vertices satisfy the problem constraints. Fig.\ref{fig:terminal_set} shows the outer polyhedral approximation of the ellipsoid terminal set.

\section{Numerical simulations}

\begin{figure}
    \centering
    \includegraphics[scale=0.3]{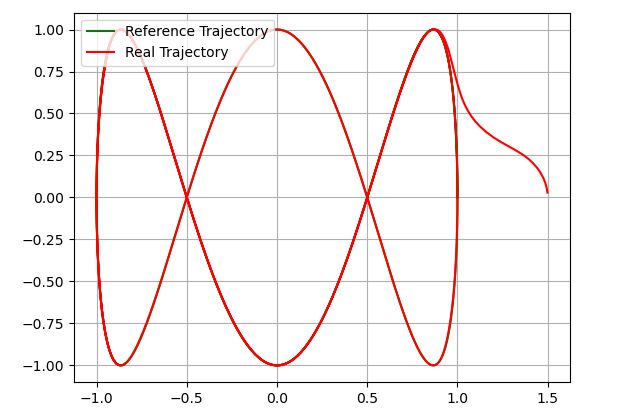}
    \caption{Tracking: robot starts from an arbitrary position and converges to the reference trajectory.}
    \label{fig:tracking}
\end{figure}

In this section, we show quantitatively the simulation results of our MPC controller in tracking a time-varying trajectory and show qualitatively how our proposed methods perform in both static and dynamic obstacle avoidance tasks. In tracking task, we use a sinusoidal trajectory (in Fig.\ref{fig:tracking} as reference), while in obstacle avoidance case we use line trajectories with constant reference velocity.

\subsection{Tracking}

\subsubsection{Horizon length (with and without terminal cost)}

First, we compare the influence of changing the length of the planning horizon $N$ on the trajectory tracking performance. In our analysis we only consider the tracking error in the X-Y axes, since a heading (i.e. $\theta$) tracking error is acceptable as long as the trajectory is followed closely. We use sum of absolute values of the errors on x- and y-axes as our metric. 

\begin{figure}
    \centering
    \includegraphics[scale=0.04]{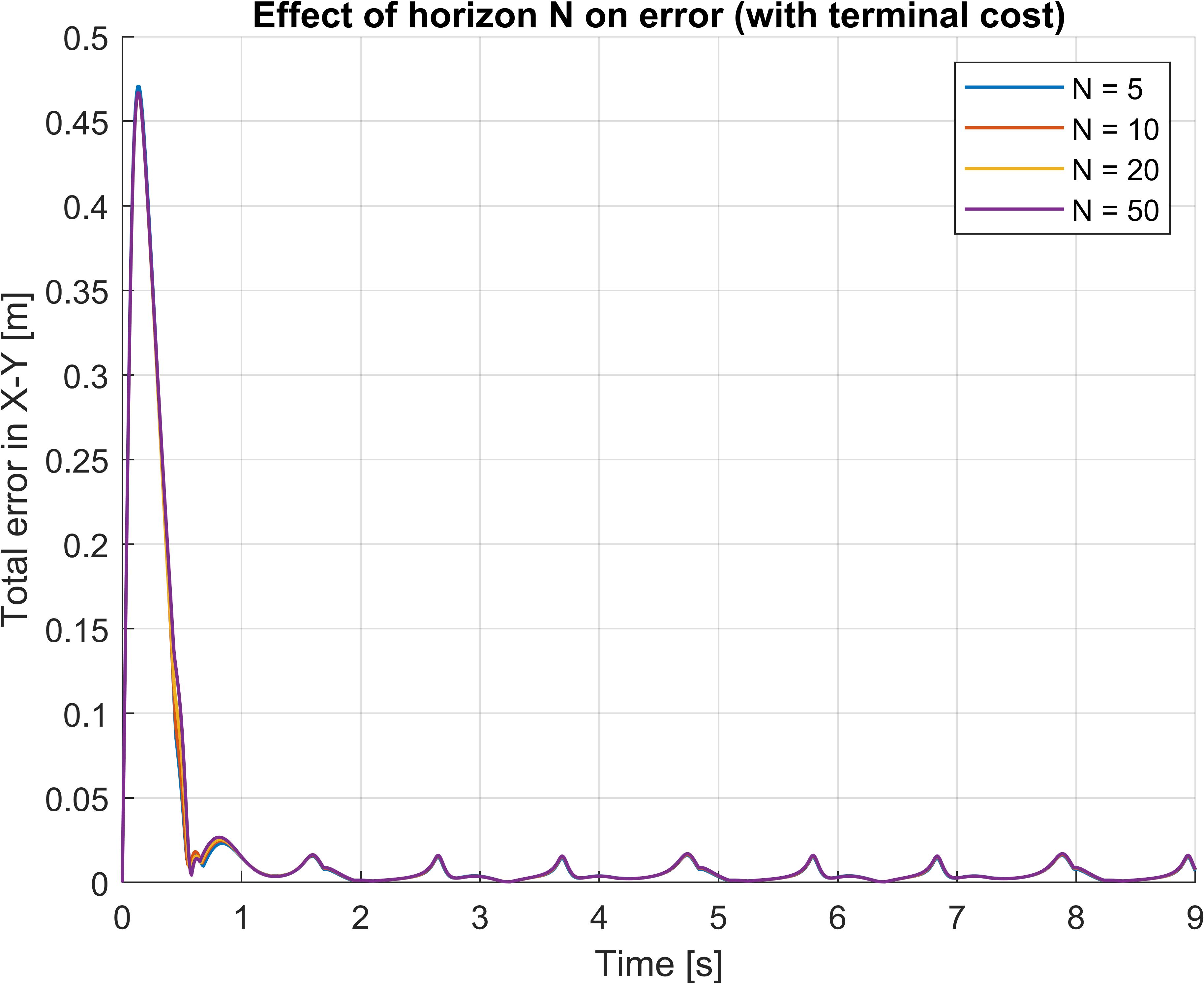}
    \caption{Illustrating how increasing the MPC horizon N affects the tracking performance.} 
    \label{fig:N_error_terminal}
\end{figure}

\begin{figure}\label{fig:N2}
    \centering
    \includegraphics[scale=0.04]{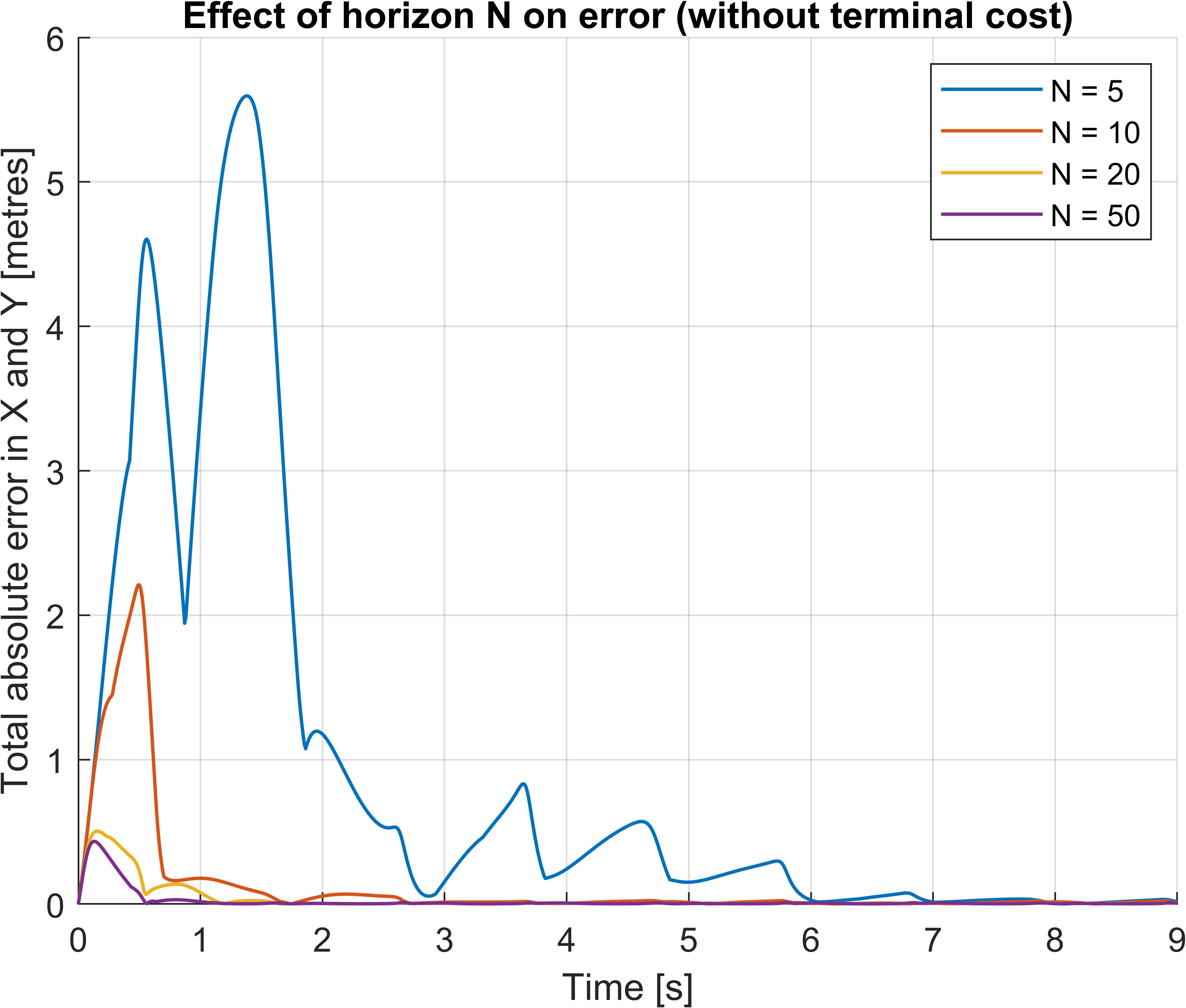}
    \caption{Without a terminal cost the tracking performance highly degrades for low ($N\leq 20$) horizon length. As the horizon increases the tracking error begins to converge to the one obtained by the terminal cost controller, shown in Fig. \ref{fig:N_error_terminal}.}
    \label{fig:N_error_no_terminal}
\end{figure}

It is worth mentioning that in our experiment, in order to make the computation more efficient, instead of using terminal set constraint, we increase the terminal cost $V_f(x,i)$ by a parameter $\beta (\beta \geq 1)$. By correctly choosing the parameter, it can have similar effect as using the terminal set constraint. Fig. \ref{fig:N_error_terminal} and Fig. \ref{fig:N_error_no_terminal} show the effect of horizon length $N$ on the MPC controller with or without terminal cost $V_f(x, i)$.

As can be observed from Fig. \ref{fig:N_error_terminal}, with using the terminal cost $V_f(x, i)$, the horizon length $N$ has minor effect on the tracking errors. Interestingly, using the same state and input costs but without a terminal cost, as can be seen in Fig. \ref{fig:N_error_no_terminal}, leads to a much worse trajectory tracking for small horizons and increasing the horizon length $N$ from 5 to 20 does have significant effect on the tracking performance. 

Our illustration is that through adding an approximation of optimal cost-to-go with infinite horizon as terminal cost, the optimal solution from the MPC problem is less 'short-sighted' and tends to steer the states towards the temporary origin faster with higher control inputs. By adding a soft constraint on the terminal states, the tracking errors converge better. And since we approximate the infinite horizon optimal cost-to-go, the stage costs of the $N$ steps constitute less in the total cost and therefore have less effect. 

In contrast, without terminal cost penalty, the MPC controller becomes more 'short-sighted' and therefore horizon length $N$ has considerable effect. That also explains above certain horizon length N, the performance tends to converge as well. ($N$ increases from 20 to 50)\\

\subsubsection{Terminal cost}

\begin{figure}
    \centering
    \includegraphics[scale=0.06]{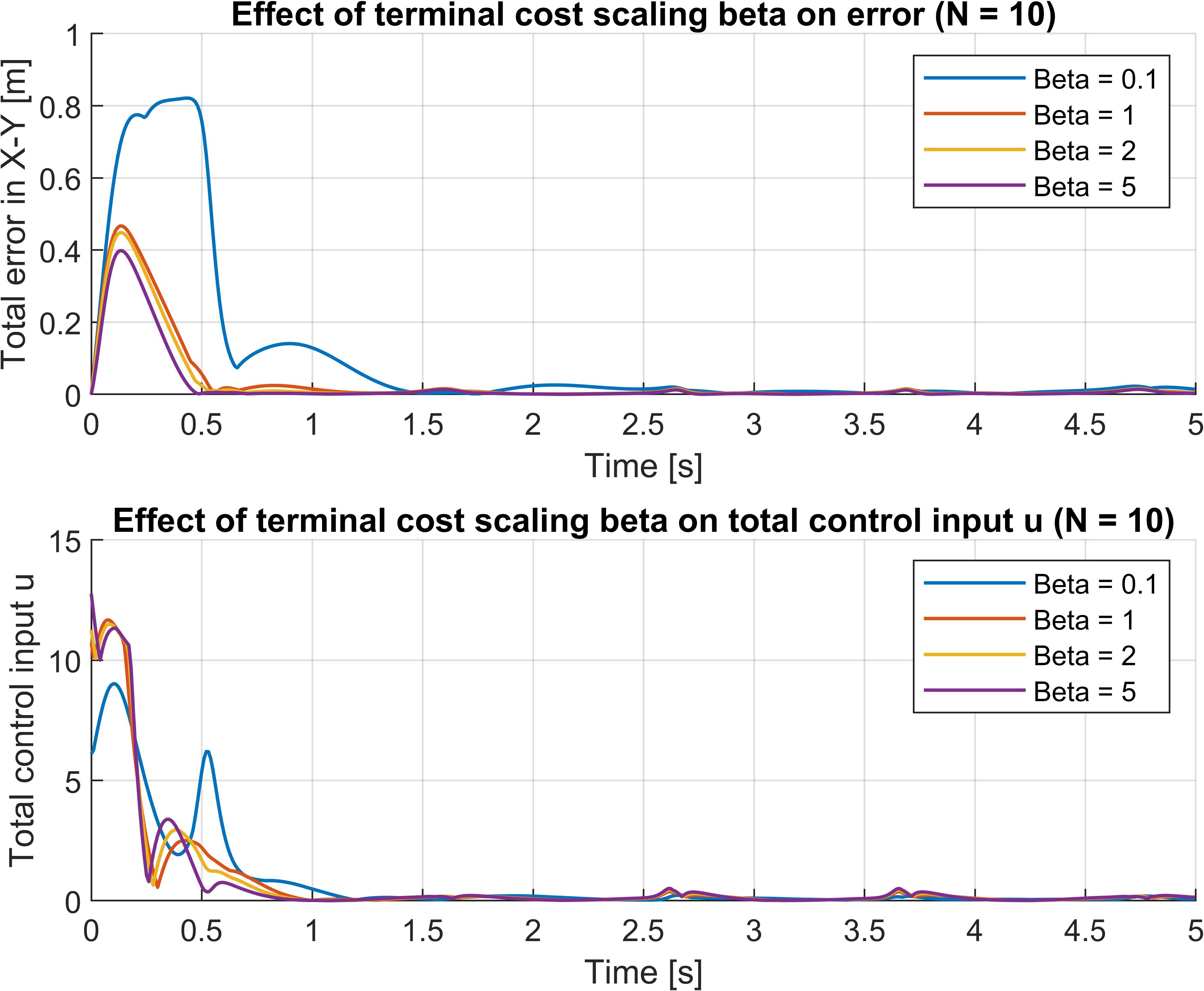}
    \caption{Showing how the $\beta$ scaling of the terminal cost changes the tracking error and control input respectively. As $\beta$ increases, they both converge.}
    \label{fig:beta_terminal_cost}
\end{figure}

Next, we analyse the effect of increasing the terminal cost by a scaling factor $\beta$ ($V_f(x)= \frac{1}{2}\beta x^TPx, \beta \geq 0$) for a constant horizon ($N=10$) on the tracking performance. The two plots in Fig. \ref{fig:beta_terminal_cost} show how the total tracking errors and the total control inputs change, respectively.

As our design choice of terminal set is a sublevel set of terminal cost $V_f(x, i)$, which means the shape of the terminal set $\mathbb{X}_f(i)$ is exactly the shape of the contour lines of the terminal cost $V_f(x,i)$. This is also the reason that we could replace the hard terminal constraint by adding the terminal cost by a scaling factor $\beta$ ($\beta \geq 1$). By reasonably choosing the $\beta$, the actual terminal states can be steered to and remain in the sublevel set as well, with decreasing the terminal cost. 

As is shown in the figure above, except for the setting of $\beta = 0.1$ (blue lines in the plots), all the $\beta$ that greater than 1 have highly similar effect on both tracking errors and control inputs ($\beta=5$ yields slightly lower errors and higher control inputs), meaning that the solution of MPC problem converges to the ideal terminal set. The setting $\beta = 0.1$ resembles the setting without terminal cost above. Illustration above also holds here.\\

\subsubsection{Unconstrained LQR vs MPC}

In this section, we compare the tracking performance and the related control inputs of our MPC with the controller that uses LQR control gain $K(i)$ computed at each time step. As this LQR setting is without control input constraints, it can exert arbitrarily large control inputs to yield lower errors.

\begin{figure}
    \centering
    \includegraphics[scale=0.06]{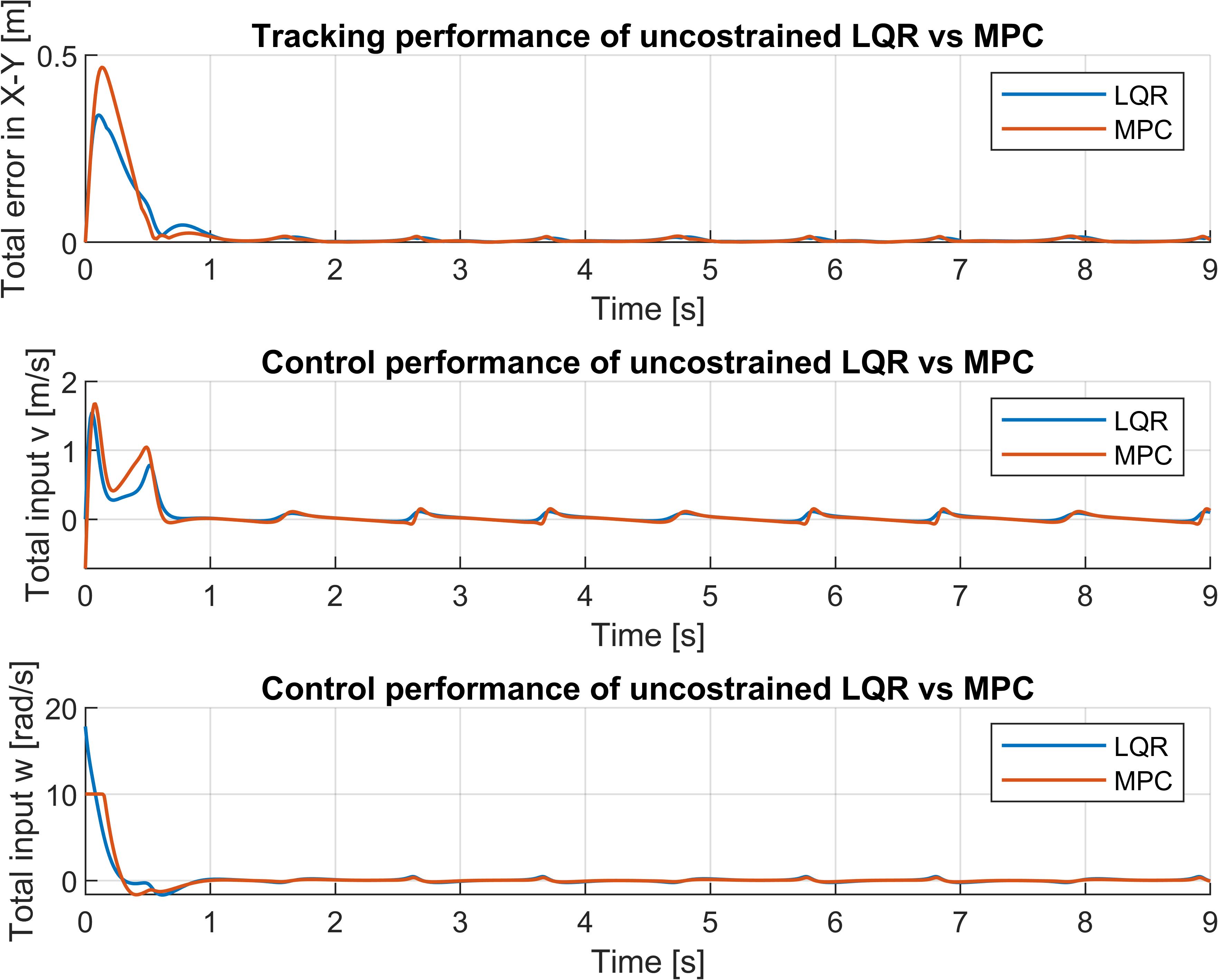}
    \caption{Three plots depicting the comparative performance in terms of tracking error and control inputs (linear velocity \textit{v} and angular velocity $\omega$) for the unconstrained LQR and the MPC controller ($N$=10).}
    \label{fig:LQR_vs_MPC}
\end{figure}

The plots above show that at the beginning of the simulation the LQR controller uses higher control inputs to yield slightly lower tracking errors. As is shown clearly, the control input of angular velocity $\omega$ is upper bounded by 10 for the MPC controller. The LQR controller, on the other hand, is unconstrained and, as the plot shows, actually exceeds the input constraints (the initial angular velocity is around 18 rad/s, and the constraint is at 10 rad/s). 
After around 1 second the results given by the two controllers mostly coincide with each other. This also validates our argument in the proof of the asymptotic stability, around the origin, that by \textit{at least} using LQR control gain, our system can yield local Lyapunov decrease. Here we qualitatively validate that when tracking errors converge to the origin, the control input given by our MPC controller also converges to the LQR controller, which is feasible around the origin. \\

\subsubsection{Lyapunov decrease}

\begin{figure}
    \centering
    \includegraphics[scale=0.05]{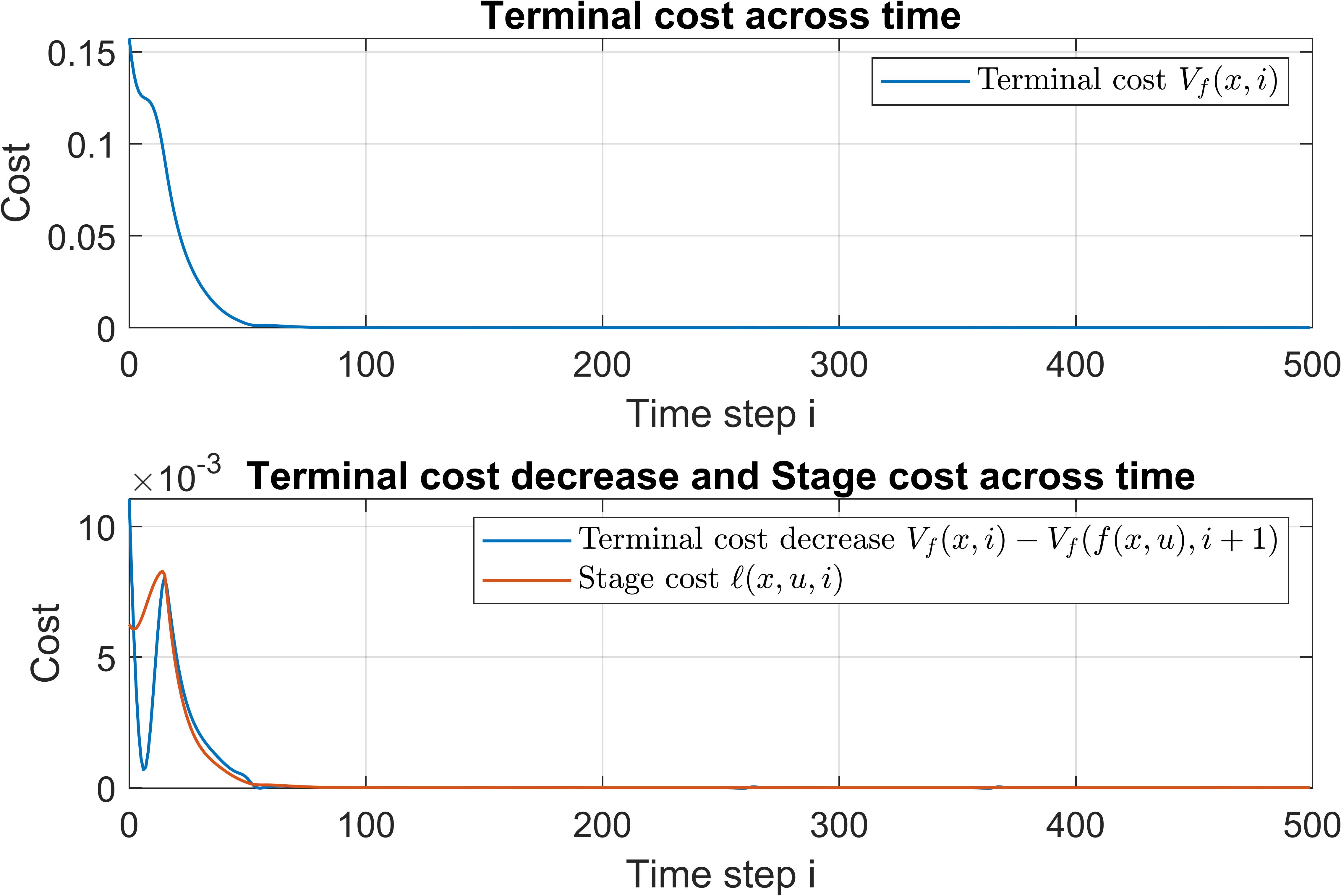}
    \caption{A-posteriori confirming the assumption on the positive definiteness of the stage cost and the terminal cost. Furthermore, experimentally showing the Lyapunov decrease of the terminal cost.}
    \label{fig:terminal_cost_posteriori}
\end{figure}

We have analytically proved the Lyapunov decrease, here we experimentally verified the satisfaction of the Assumption 2.33(a). As can be seen in the top plot on Fig. \ref{fig:terminal_cost_posteriori}, the terminal cost is monotonically decreasing until it reaches 0 once the robot is following the trajectory closely. The bottom plot confirms Assumption 2.33(a) in that once the states get close enough to the origin, the Lyapunov decrease between stages is greater or equal to the stage cost: $V_f(x,i)-V_f(f(x,u),i+1) \geq \ell(x,u,i)$.

\subsection{Static obstacle}

\begin{figure}
    \centering
    \includegraphics[scale=0.35]{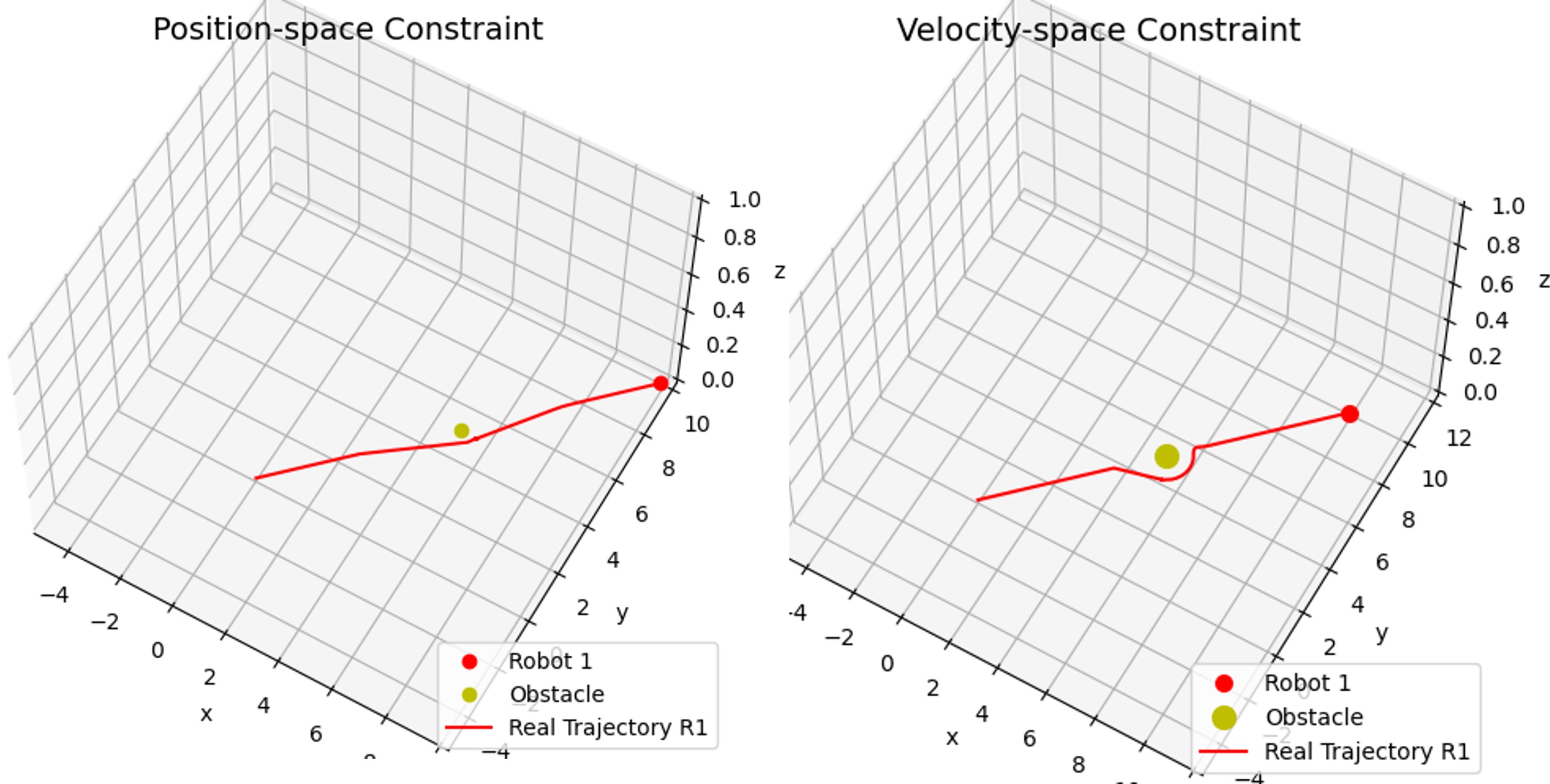}
    \caption{Comparison between the two obstacle avoidance methods when considering a static obstacle in the centre of the trajectory.}
    \label{fig:static}
\end{figure}

Figure \ref{fig:static} shows the simulation results of our MPC controller with \textit{linear} constraints in position (left) and velocity space (right) when avoiding a static obstacle on the reference path. This scenario is common in the field of robotics, where a global path given by the path planner (e.g. RRT method \cite{RRT}) is computed offline, as it is computationally expensive. If a new static or dynamic obstacle is present on the reference path the MPC controller can locally re-adjust the trajectory, without having to re-run the expensive global path planner. 

As can be seen from figure \ref{fig:static}, both of our methods give a smooth trajectory while avoiding the local obstacle. In our experiments, we also found out that because of the non-holonomic constraints on the system kinematics, a 90 degree linear constraint (rotates with movement of the robot) between the obstacle and the robot can result in infeasibility of the MPC problem. Adding slack variables to loose the constraint can help but loses the smoothness of the trajectory and collision avoidance guarantee. 

By contrast, our proposed linearized constraint in velocity space explicitly models the non-holonomic property of the system and therefore avoids infeasible optimization problems. As the simulation shows, it gives smooth circular trajectory while avoiding the obstacle. We tested our algorithm for multiple safety radii, all of which managed to avoid the obstacle smoothly and efficiently.

\subsection{Dynamic obstacle}

For dynamic obstacle avoidance we tested our algorithm in the scenarios where two unicycle robots have conflicting global reference paths, either face-to-face or with a point of intersection. For a clearer illustration of our results, we only define collision avoidance constraint for one of them, but note that our method can be generalised to multiple robots cooperatively by planning in their relative velocity spaces. 

\begin{figure}
    \centering
    \includegraphics[scale=0.3]{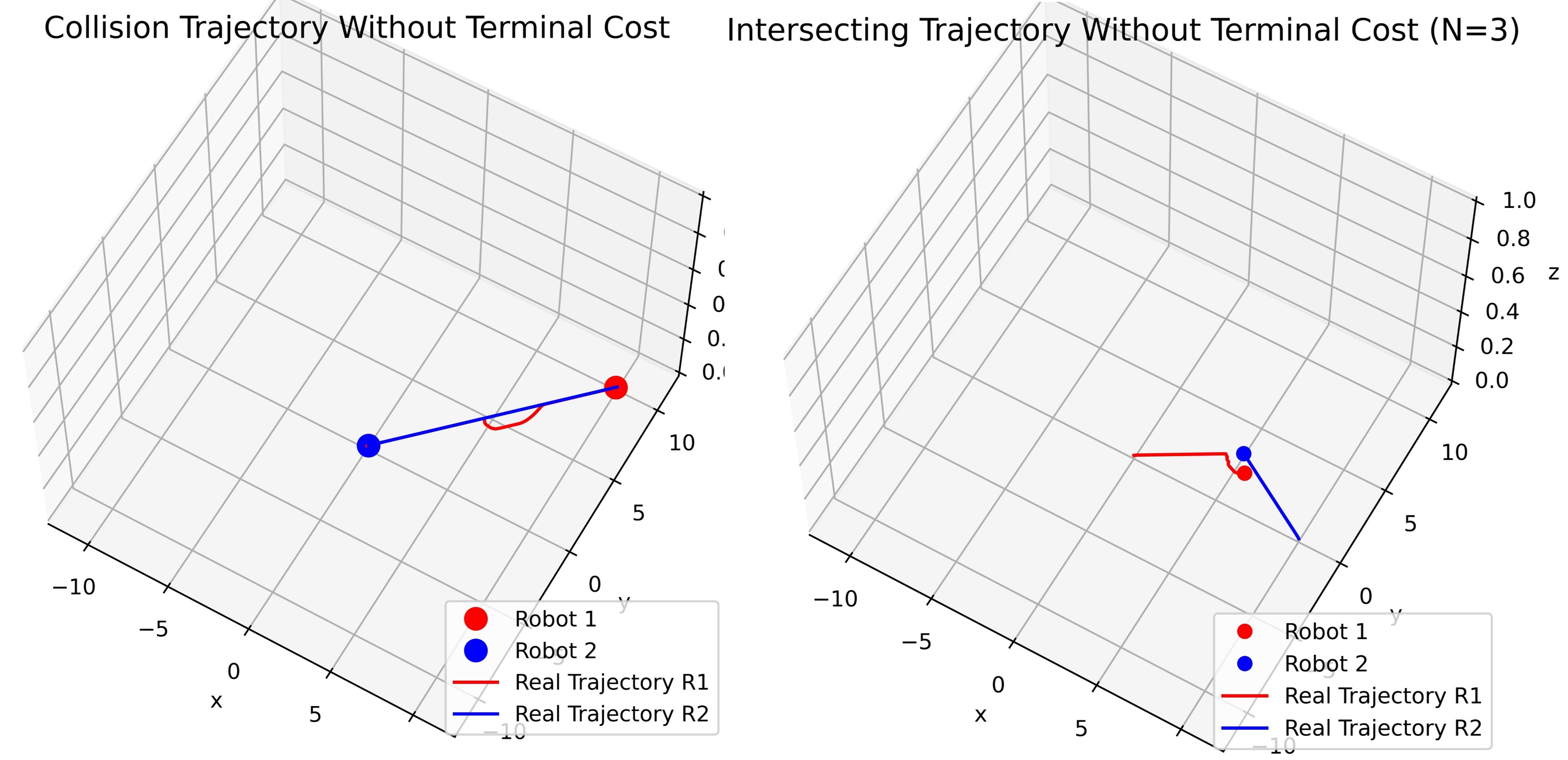}
    \caption{Qualitative analysis of the collision avoidance between two robots (with only Robot 1 planning to avoid collision), without a terminal cost and for a time horizon of 3.}
    \label{fig:dynamic_no_Vf}
\end{figure}

\begin{figure}
    \centering
    \includegraphics[scale=0.4]{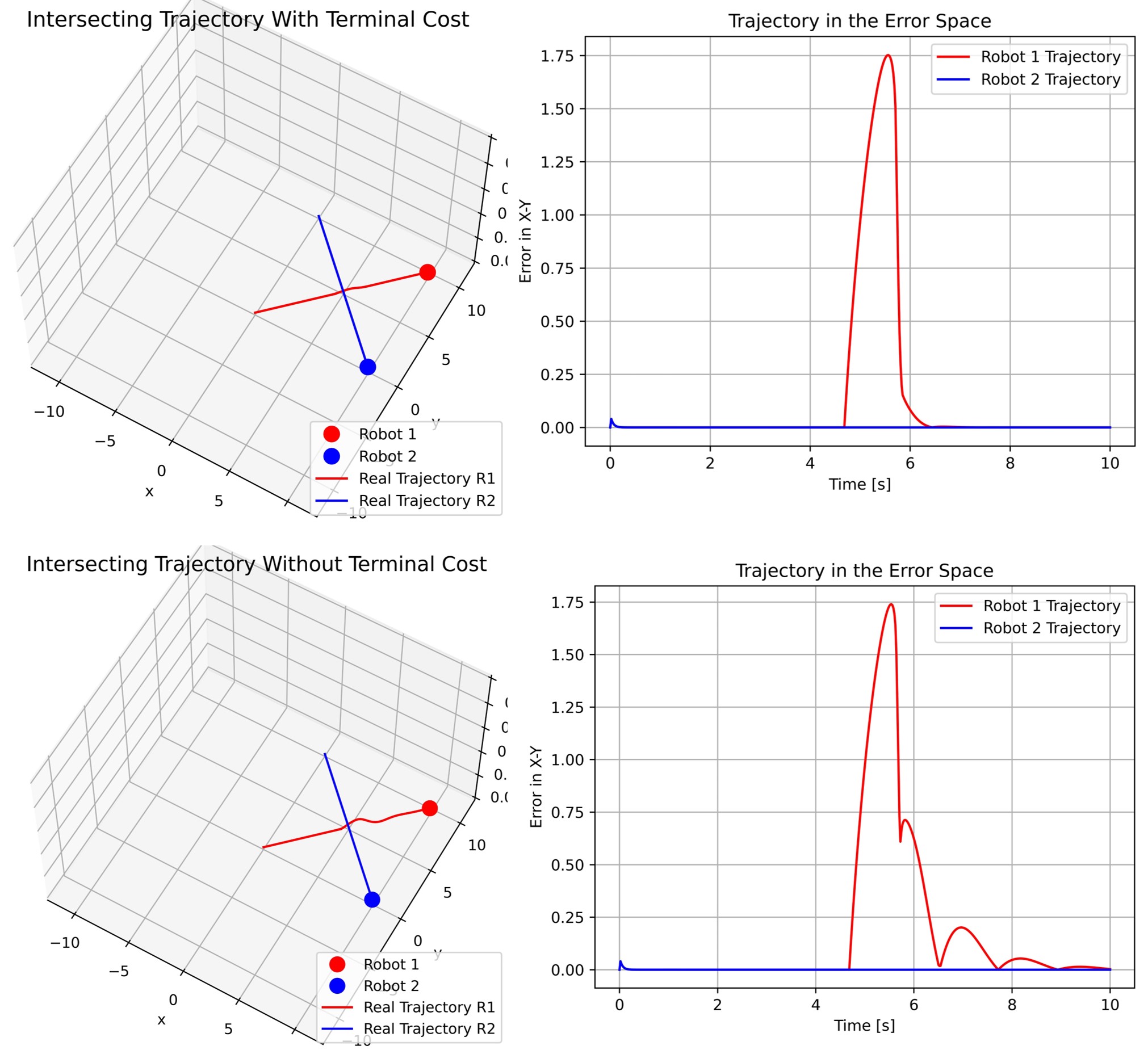}
    \caption{Comparison of the performance with and without a terminal cost for an intersecting trajectory. Notice that Robot 1 tends to deviate less from the reference trajectory when adding a terminal cost.}
    \label{fig:dynamic_with_Vf}
\end{figure}




As the simulation results in Fig. \ref{fig:dynamic_no_Vf} show, in the scenario where two robots try to exchange their positions (facing each other), the first robot can deviate from its reference path and converge back after avoiding the second robot. The resulting trajectory is comparably smooth when compared with the solution given by state-of-the-art NLP solvers\cite{NLP}. In the second scenario (Fig. \ref{fig:dynamic_no_Vf}, right) the robot also successfully adjusts its trajectory to avoid the other robot at the point of intersection between the trajectories.\\

Figure \ref{fig:dynamic_with_Vf} shows how the addition of a terminal cost affects the quality of the trajectory on an intersecting path. It is important to mention that the first robot slows down in order to avoid the second robot, rather than change its path, as can be observed from the trajectory plot in the error space. Without a terminal cost penalty the robot tends to deviate from the reference trajectory more and the trajectory is less smooth. On the other hand, the terminal cost (top plot) produces a much smoother final trajectory.
\footnote{For more experimental results and simulation videos please refer to our \href{https://youtu.be/nYDxWkKvzZ8}{channel}}

\section{Conclusion \& future work}

In this paper, we introduced an MPC approach for motion planning and control of non-holonomic mobile robots. Furthermore, we proposed a novel obstacle avoidance method that defines constraints in velocity space and explicitly integrates non-holonomic constraints in it. Simulation results show that our methods can generate highly smooth trajectories for local obstacle avoidance. This MPC approach can navigate non-holonomic mobile robots in dynamic environment in a computationally efficient way that only employs quadratic programming. 

We also analytically showed the recursive local asymptotic stability of our MPC controller in tracking a reference trajectory without local obstacle avoidance. We proved stability for a time-invariant trajectory and then expanded this proof for a time-varying system. We used extensive experimental results to analyse and validate our design choices of the MPC controller. 

For future work, on the one hand, we want to generalize our method in multiple robots coordination case. Planning in their relative velocity space is a promising direction, and centralised planning for multiple robots can potentially yield smoother and better trajectories. On the other hand, for theoretic stability analysis, we plan to employ an iterative process that resembles iLQR \cite{iLQR} so that our linearisation assumptions still hold when the robot is locally avoiding the obstacle and deviating form the reference trajectory.


\bibliographystyle{IEEEtran}

\bibliography{vo_mpc}

\newpage

\section{Appendix}

\subsection{Derivation of system dynamics linearization}

A unicycle model is given by:

\small
\begin{equation}\label{model}
    \left(
    \begin{matrix}
    \dot x(t)\\
    \dot y(t)\\
    \dot \theta(t)\\
    \end{matrix}
    \right)
    = \dot z(t) = f(z(t), u(t)) = 
    \left(
    \begin{matrix}
    v(t)cos(\theta(t))\\
    v(t)sin(\theta(t))\\
    \omega(t) \\
    \end{matrix}
    \right)
\end{equation}
\normalsize

Given a reference path $r = [x_r, y_r]$, we can derive the reference control input by using Eq.\eqref{model}, which gives (for simplifying the notations, we ignore the time variation here):
\small
\begin{equation}
    \begin{split}
    v_r &= \sqrt{\dot x_r^2(t) + \dot y_r^2(t)}\\
    \theta_r &= atan2(\dot y_r(t), \dot x_r(t))\\   
    \omega_r &= \frac{\dot x_r(t)\ddot y_r(t) - \dot y_r(t) \ddot x_r(t)}{\dot x_r^2(t) + y_r^2(t)}
    \end{split}
\end{equation}
\normalsize

For MPC regulation problem, we define error vector in robot's \textit{local frame} as new state $e(t) = [e_1(t), e_2(t), e_3(t)]$:

\small
\begin{equation}
    \begin{split}
    e(t) &= 
    \left[
    \begin{matrix}
    cos\theta & sin\theta & 0\\
    -sin\theta & cos\theta & 0\\
    0 & 0 & 1\\
    \end{matrix}
    \right]
    \left[
    \begin{matrix}
    x_r - x\\
    y_r - y\\
    \theta_r - \theta\\
    \end{matrix}
    \right]\\
    &=
    \left[
    \begin{matrix}
    (x_r-x)cos\theta + (y_r-y)sin\theta\\
    (x - x_r)sin\theta + (y_r - y)cos\theta\\
    \theta_r - \theta\\
    \end{matrix}
    \right]
    \end{split}
\end{equation}
\normalsize

By taking the gradient to the error, we have the error dynamics:

\scriptsize{
\begin{equation}\label{error}
    \dot e = 
    \left[
    \begin{matrix}
    (\dot x_r - \dot x)cos\theta - \dot \theta (x_r - x)sin\theta + (\dot y_r - \dot y)sin\theta + \dot \theta (y_r - y)cos \theta\\
    \dot \theta(x-x_r)cos\theta + (\dot x-\dot x_r)sin\theta + (\dot y_r - \dot y)cos\theta + \dot \theta(y-y_r)sin\theta\\
    \dot e_3\\
    \end{matrix}
    \right]
\end{equation}}

For the first element of the equation above, we observed that:

\begin{equation*}
        \dot \theta (y_r - y)cos \theta - \dot \theta (x_r - x)sin\theta  = e_2\omega
\end{equation*}

\begin{equation*}
        \begin{split}
            (\dot x_r - \dot x)cos\theta + (\dot y_r - \dot y)sin\theta &= (v_r cos\theta - v cos_\theta)cos\theta + (v_r sin\theta_r - v sin\theta)sin\theta \\
            &= v_r cos(\theta_r - \theta) - v\\
            &= v_r cos(e_3) - v
        \end{split}
\end{equation*}

By using the equivalent substitution to the second element of \eqref{error}, we have:
\small
\begin{equation}
    \dot e = 
    \left[
    \begin{matrix}
    v_r cos(e_3) - v + e_2\omega\\
    v_r sin(e_3) - e_1 \omega\\
    \omega_r - \omega\\
    \end{matrix}
    \right]
\end{equation}
\normalsize
By using the control input $u_f = [v_r, w_r]^T$ as feed-forward control and assuming our MPC regulator operates around the reference ($cos(e_3) \approx 1$), we yield:
\small
\begin{equation}
    \begin{split}
    \dot e & \approx 
    \left[
    \begin{matrix}
    v_r - (v_r + v_b) + e_2(\omega_r + \omega_b)\\
    v_r sin(e_3) - e_1 (\omega_r + \omega_b)\\
    \omega_r - (\omega_r + \omega_b)\\
    \end{matrix}
    \right]\\
    &=
    \left[
    \begin{matrix}
    e_2\omega_r\\
    v_r sin(e_3) - e_1\omega_r\\
    0\\
    \end{matrix}
    \right] +  
    \left[
    \begin{matrix}
    -1 & e_2\\
    0 & -e_1\\
    0 & -1\\
    \end{matrix}
    \right]u_b
    \end{split}
\end{equation}
\normalsize
Notice that point $[e(t), u_b(t)]^T = [0_3, 0_2]^T$ is an equilibrium, where we can use first-order Taylor approximation to linearize the error dynamics:
\small
\begin{equation}
    \dot e =
    \left[
    \begin{matrix}
    0 & \omega_r & 0\\
    - \omega_r & 0 & v_r\\
    0 & 0 & 0\\
    \end{matrix}
    \right]e +  
    \left[
    \begin{matrix}
    -1 & 0\\
    0 & 0\\
    0 & -1\\
    \end{matrix}
    \right]u_b
\end{equation}
\normalsize
By using forward Euler approximation with sampling time of T, we have the discrete-time error dynamics, which can be used for our MPC regulator:
\small
\begin{equation}
    e(k+1) =
    \left[
    \begin{matrix}
    1 & T \omega_r & 0\\
    - T \omega_r & 0 & T v_r\\
    0 & 0 & 1\\
    \end{matrix}
    \right]e(k) +  
    \left[
    \begin{matrix}
    -T & 0\\
    0 & 0\\
    0 & -T\\
    \end{matrix}
    \right]u_b(k)
\end{equation}
\normalsize
\subsection{Derivation of velocity obstacle}

This derivation is based on \cite{VO}

Given two robots $i$ and $j$, the velocity obstacle can be described as follows:
\small
\begin{equation*}
    VO = \{u_i \text{for which} \|(p_i+tu_i)-(p_j+tu_j) \leq r_i + r_j \text{for some} t \in [0,\tau]\}
\end{equation*}
\normalsize
which means:
\small
\begin{equation*}
    \begin{split}
        &\|p_i + tu_i - p_j -tu_j\| \leq r_i + r_j\\
        &\|\frac{p_i-p_j}{t} + u_i - u_j\| \leq \frac{r_i+r_j}{t}\\
        &\|u_i - (u_j+\frac{p_j-p_i}{t})\| \leq \frac{r_i+r_j}{t}
    \end{split}
\end{equation*}
\normalsize
Therefore, the velocity obstacle in robot $i$'s velocity space is:
\small
\begin{equation*}
    u_i \in D(\frac{p_j-p_i}{t}, \frac{r_i+r_j}{t})
\end{equation*}
\normalsize
That is: $VO^{\tau} = \mathop{\cup} \limits_{0 \leq t \leq \tau} D(\frac{p_j - p_i}{t}+u_j, \frac{r_i+r_j}{t})
$

\end{document}